\pdfoutput=1

\documentclass[11pt]{article}

\usepackage[preprint]{acl}

\usepackage{times}
\usepackage{latexsym}

\usepackage[T1]{fontenc}

\usepackage[utf8]{inputenc}

\usepackage{microtype}

\usepackage{inconsolata}

\usepackage{graphicx}

\usepackage{amssymb}
\usepackage{booktabs} 
\usepackage{graphicx} 
\usepackage{adjustbox} 
\usepackage{subcaption}

\title{MOOM: Maintenance, Organization and Optimization of Memory in Ultra-Long Role-Playing Dialogues}

\author{
 \textbf{Weishu Chen\textsuperscript{1}},
 \textbf{Jinyi Tang\textsuperscript{2}},
 \textbf{Zhouhui Hou\textsuperscript{2}},
 \textbf{Shihao Han\textsuperscript{2}},
 \textbf{Mingjie Zhan\textsuperscript{2}},
 \textbf{Zhiyuan Huang\textsuperscript{2}},
\\ 
 \textbf{Delong Liu\textsuperscript{1}},
 \textbf{Jiawei Guo \textsuperscript{2}},
 \textbf{Zhicheng Zhao\textsuperscript{1,3}},
 \textbf{Fei Su\textsuperscript{1,3}},
\\
\\
 \textsuperscript{1}Beijing University of Posts and Telecommunications,
 \textsuperscript{2}SenseTime,\\
 \textsuperscript{3}Beijing Key Laboratory of Network System and Network Culture,\\
 \small{
   chenweishu@bupt.edu.cn
 } 
}

\begin{document}
\maketitle
\begin{abstract}
Memory extraction is crucial for maintaining coherent ultra-long dialogues in human-robot role-playing scenarios. However, existing methods often exhibit uncontrolled memory growth. To address this, we propose MOOM, the first dual-branch memory plugin that leverages literary theory by modeling plot development and character portrayal as core storytelling elements. Specifically, one branch summarizes plot conflicts across multiple time scales, while the other extracts the user's character profile. MOOM further integrates a forgetting mechanism, inspired by the ``competition-inhibition'' memory theory, to constrain memory capacity and mitigate uncontrolled growth. Furthermore, we present ZH-4O, a Chinese ultra-long dialogue dataset specifically designed for role-playing, featuring dialogues that average 600 turns and include manually annotated memory information. Experimental results demonstrate that MOOM outperforms all state-of-the-art memory extraction methods, requiring fewer large language model invocations while maintaining a controllable memory capacity. Code is available at \href{https://github.com/cows21/MOOM-Roleplay-Dialogue}{here}.
\end{abstract}

\section{Introduction}

Large language models (LLMs) \cite{qwenreport,gpt4report, lu2024deepseek} are no longer simply regarded as practical assistants, but have been increasingly sought for obtaining emotional support. For example, users reshape the LLMs into their favorite idol, ideal companion, gaming partner, or confidant for personal concerns. This practice of assigning specific personas or characters to a language model is commonly referred to as role-playing \cite{rolepaly_oscars}, where both the language model and the users actively play their respective roles, including but not limited to themselves, fictional characters, or entirely new protagonists. However, when these narratives become excessively long, due to limited context windows or inherent issues such as hallucinations \cite{hallucination_inevitable}, current LLMs often overlook historical information provided by users or confuse critical details, potentially degrading the user experience.

In this paper, we view role-playing dialogues with LLMs as ``stories'' co-created by users with the assistance of LLMs. Traditional literary theory \cite{aspects_of_novels} identifies the plot and characters as the core elements of a story. Inspired by this, we propose the MOOM plugin to extract key information from role-playing dialogues, enabling dialogue models to access and align with the plot and character context, thereby generating responses that conform to the narrative and user-defined roles. MOOM consists of two branches: one branch summarizes the progression of the plot across varying temporal scales, while the other uses predefined keywords to construct and update character profiles. Additionally, to prevent unrestricted memory growth in ultra-long dialogues, we design a forgetting algorithm based on the ``competition-inhibition'' theory \cite{rethinking_memory}, which mimics human memory by prioritizing relevant information through temporal decay and retrieval reinforcement while suppressing outdated or noisy data. This allows MOOM to maintain flexible memory capacity while preserving a diverse range of memory information.

To evaluate the effectiveness of the MOOM plugin, we establish a new Chinese R\textbf{O}le-Playing L\textbf{O}ng Dial\textbf{O}gue Mem\textbf{O}ry Dataset named ZH-4O. First, we collect role-playing dialogues between annotators and LLMs. Subsequently, a group of independent annotators annotate key information and its corresponding positions within these dialogues. Finally, a total of 28 dialogues with memory annotations are obtained, each of which has an average length of 600 turns. Additionally, to assess the impact of memory information generated by different memory frameworks on the LLM responses, we design multiple sets of probing questions, aiming to comprehensively cover various types of role-playing information.

Our main contributions can be summarized as follows:
\begin{itemize}
    \item[(1)]
    We propose MOOM, a novel memory extraction framework that models two core story elements: ``plot development'' and ``character portrayal''. To the best of our knowledge, this is the first long-memory extraction method designed around ``story elements''.
    \item[(2)]
    We introduce a forgetting algorithm to constrain memory capacity and mitigate uncontrolled memory expansion.
    \item[(3)]
    We construct ZH-4O, a new dataset comprising authentic, ultra-long human-LLM dialogues in role-playing scenarios, along with manually annotated memory information, serving as a benchmark for evaluating memory extraction methods.
    \item[(4)]
    Experiments demonstrate that, compared with state-of-the-art methods, MOOM achieves higher memory extraction accuracy and enables LLMs to generate more efficient and contextually coherent responses.
\end{itemize}

\section{Related Work}

\subsection{Long-term Memory Extraction in Conversations}
Due to the inefficiency of using the entire dialogue history as long-term memory, several techniques have been developed to extract and organize key information from past interactions. MemoChat divides conversations into distinct topics and generates concise summaries for each sub-dialogue to construct memory \cite{lu2023memochat}. COMEDY employs a fine-tuned model to compress memory into three categories: user profiles, relationship descriptions, and event records \cite{chen2024compresscomedy}. LTM employs a personality extractor to classify users into predefined personality categories \cite{xu2022longtime}. Regarding memory updating, \cite{bae2022keepme} represents memory as unstructured text descriptions and utilizes a fine-tuned T5-based classification model to iteratively compare and update old and new memory entries. Similarly, the open-source framework Mem0\footnote{\href{https://github.com/mem0ai/mem0}{https://github.com/mem0ai/mem0}} employs a graph-based representation of memory and relies on LLMs to determine update operations for each memory entry. Unlike these approaches, which primarily rely on semantic similarity or topic segmentation, MOOM integrates literary theory to model narrative and persona, offering a more structured and contextually relevant memory framework.

\subsection{Forgetting in Long-term Memory Management}
Managing long-term memory in conversations involves strategies to control memory capacity and update or discard information over time to maintain efficiency. MemoryBank incorporates the Ebbinghaus Forgetting Curve to model memory decay over time \cite{memorybank}. MemoCRS utilizes a First-In-First-Out (FIFO) strategy to control memory capacity \cite{memocrsforget}. In general, most memory extraction methods rely on semantic similarity comparisons between historical and new memory entries during the update process to decide whether to retain or discard specific information \cite{qian2024chatdevupdate1, schuurmans2023memoryupdate2, bae2022keepme}. However, existing methods still struggle with uncontrolled memory growth in ultra-long dialogues. In contrast, we propose a competition-inhibition-based forgetting algorithm to mitigate uncontrolled memory expansion and improve memory efficiency.

\begin{figure*}[h]
    \centering
    \includegraphics[width=0.9\textwidth]{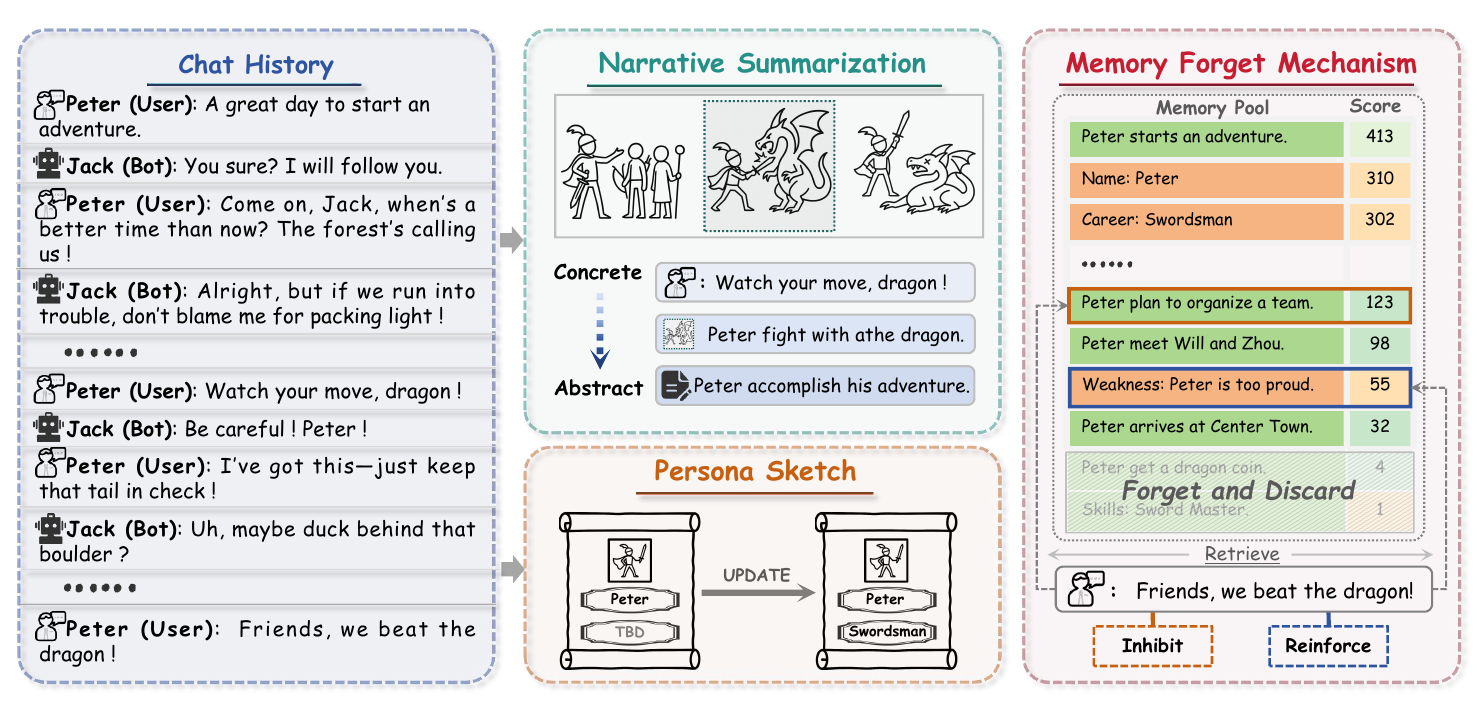}
    \caption{Overview of the MOOM framework.}
    \label{fig:main_method}
\end{figure*}

\subsection{Dialogue Dataset}
Numerous efforts have been made to construct dialogue datasets \cite{dataset1_shao2023character, dataset2_li2023chatharuhi, dataset3_wang-etal-2024-rolellm}. 
For instance, some datasets, such as those in \cite{xu2022longtime, zhou2023characterglmdataset1, tu2024characterevaldataset2}, consist of human-annotated dialogues but are limited by their relatively short lengths, averaging fewer than 20 turns. 
In contrast, datasets like LoCoMo, which average 300 turns \cite{maharana2024evaluatinglocomo}, are primarily machine-generated, with human intervention restricted to correcting logical errors. 
As a result, these machine-generated datasets do not fully capture the dynamics of human-machine conversations. 
Consequently, despite the variety of datasets available for conversational large language models, there remains a scarcity of datasets specifically designed for ultra-long dialogues with reliable memory annotations. 
To address these limitations, we introduce ZH-4O, an ultra-long dialogue dataset comprising real human-computer interactions, with an average of 600 turns per dialogue and manually annotated memory information.

\section{Methodology}
The task of memory extraction can be defined as follows: A dialogue dataset $D(d_1,d_2,...,d_t)$ consists of a series of prior dialogue sessions between a chatbot and a user. At a given time step $t$, the dialogue context is represented as $d_t=\{c_t,u_t\}$, where $c$ and $u$ denote the chatbot’s and the user’s responses, respectively. The objective of the memory extraction task is to leverage a model to extract memories $M=\{m_1,m_2,...,m_{t-1}\}$ from $D$ that are related to the user. Hence, an effective memory extraction framework is expected to exhibit the following characteristics: (1) content completeness, (2) low latency, and (3) controllable capacity.

The classic literary theory posits that the essence of a story lies in its narrative and characters \cite{aspects_of_novels}, which is compatible with role-playing scenarios. Therefore, we regard human-computer dialogues as stories co-created by users and chatbot first, and then propose our memory framework that comprises two branches: the first Narrative Summarization Branch (NSB) summarizes storyline conflicts across multi-dimensional time scales, while the second Persona Construction Branch (PCB) captures user personas to deliver a personalized experience. Finally, to address the issue of unbounded memory growth in ultra-long dialogues, we introduce a forgetting algorithm based on the ``competition-inhibition'' memory theory \cite{competition1_lindsay2013human, competition2_anderson1994remembering} to constrain memory quantity.

\subsection{Narrative Summarization Branch(NSB)}
\label{method:sumamry}

\begin{figure}[h]
    \centering
    \includegraphics[width=0.4\textwidth]{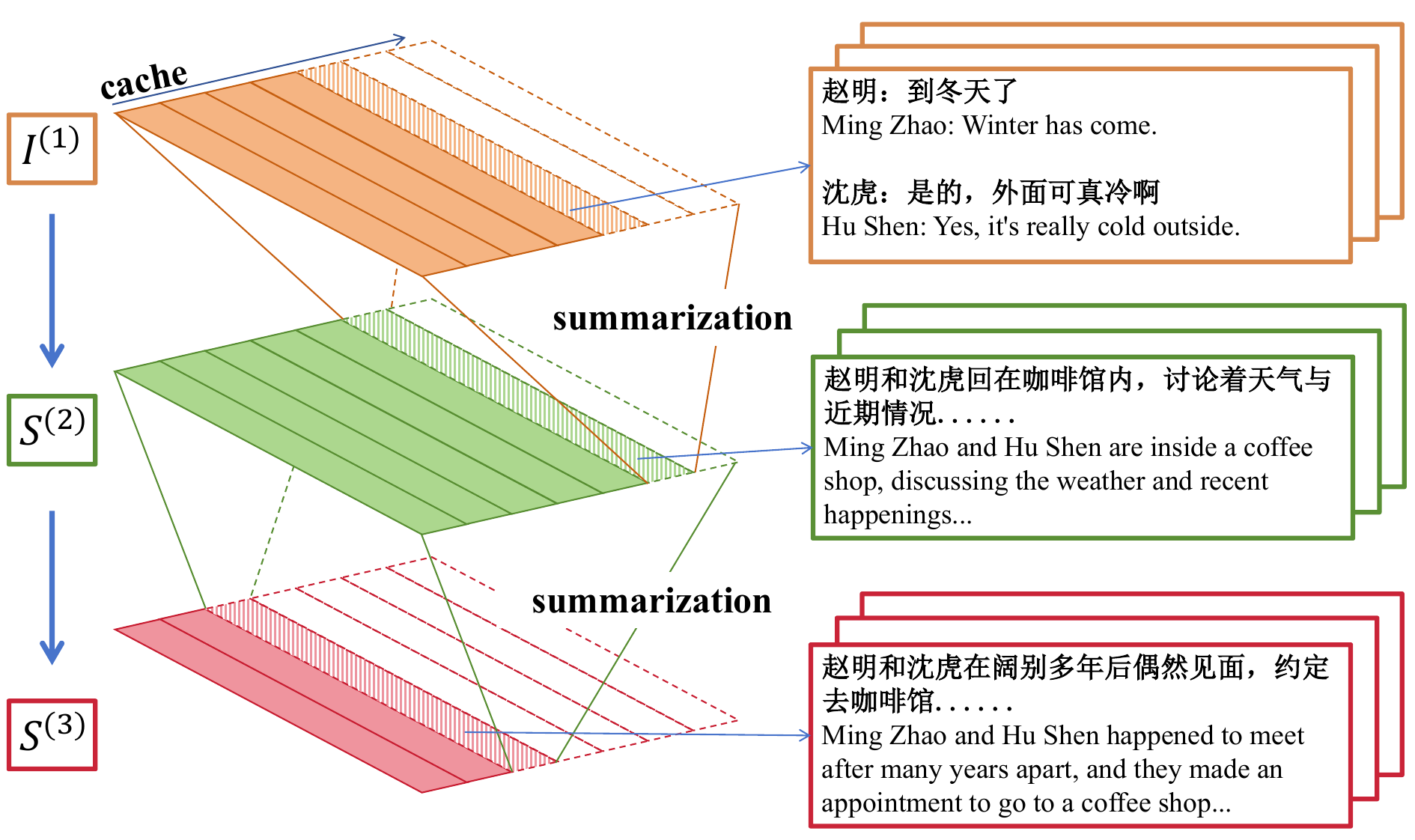}
    \caption{Illustration of the hierarchical summarization process in the Narrative Summarization Branch.
    The input dialogue $I^{(1)}$ is first condensed into $S^{(2)}$, capturing key events, and then further refined into $S^{(3)}$, which extracts the core narrative elements and focuses on high-level story progression.}
    \label{fig:summarize}
\end{figure}

In lengthy dialogues within a role-playing scenario, the narrative unfolded through a sequence of conflicts, resolutions, and turning points, which collectively define the storyline. Building on this observation, the NSB is designed to extracts key plots by capturing critical moments across multiple temporal scales. Moreover, To handle ultra-long dialogues, we implement a hierarchical summarization algorithm that structures and compresses information flow while preserving pivotal narrative elements.

Specifically, for each dialogue turn, we initially store it as raw information flow, capturing the micro-level plot development. When the cumulative dialogue turns reach the threshold $\theta_1$ for first-level packaging, they will be aggregated into a first-level information unit $I_{j}^{(1)}$, which can be expressed as:

\begin{equation}
  \label{eq:sum1}
  \scalebox{0.9}{$I_{j}^{(1)}=\{d_{(j-1)\theta_1+1},d_{(j-1)\theta_1+2},...,d_{j\theta_1}\}, j\in \mathbb{N}_{+}$}
\end{equation}

The primary units encapsulate the original dialogue into structured segments. The segments are then iteratively summarized. Whenever the number of segments at a given level reaches the packaging threshold $\theta_m$, they are forwarded to the LLM for integration and summarization, resulting in a more abstract summary:

\begin{equation}
  \label{eq:sum3}
  \scalebox{0.9}{$S_{l}^{(m+1)}=\mathrm{LLM}(\{S_{(l-1)\theta_m+1}^{(m)},...,S_{l\theta_m}^{(m)}\}), l\in \mathbb{N}_{+}$}
\end{equation}

In this work, the original packaged dialogue undergoes two iterative summarization stages, as illustrated in Fig. \ref{fig:summarize}. In our implementation, the threshold $\theta_1$ is empirically set to 6, while both $\theta_2$ and $\theta_3$ are set to 5.

This multi-level compression and abstraction process distills the dialogue into a high-density structured storyline, efficiently reducing the dependency on extended contexts while preserving the storylines' critical elements.


\subsection{Persona Construction Branch(PCB)}
\label{method:kv}

The PCB focuses on efficiently constructing and dynamically updating user personas via combining the strengths of LLMs and lightweight models. Firstly, we leverage large-scale data to extract abstract persona keys, such as ``Name'', ``Preferences'', and ``Profession'', which serve as the foundation for user characterization (details are provided in the Appendix \ref{app:personakeys}). 
Then, the LLM extracts corresponding persona values for these keys at specific intervals of dialogue turns, thereby generating a persona snapshot.

After a new persona snapshot is generated, it will be integrated with the existing persona sketch through a fusion mechanism tailored to the properties of each key. Given that leveraging LLMs for fusion is time-consuming, we reduce reliance on LLMs by introducing three distinct merging strategies, which prioritize efficiency while maintaining high-quality integration:

\begin{itemize}
    \item[(1)]
    \textbf{Rule-based Merging}: For keys with clear update rules, such as those with replaceable or appendable values, we either append new values from the persona snapshot to the corresponding existing values or overwrite old values with new ones.
    \item[(2)]
    \textbf{Embedding-based Merging}: For keys prone to contradictions, such as ``Favorite Animals'' and ``Disliked Animals'', we compute the BGE score\cite{bgescore} between conflicting values. Those outdated values with high similarity will be deleted
    \item[(3)]
    \textbf{LLM-based Merging}: For keys that cannot be efficiently handled by rules or embeddings, we delegate the merging scheme to the LL, and it will assess and integrates the new and old information.
\end{itemize}

In implementation, the majority of keys rely on rule-based merging, while embedding-based and LLM-based strategies are reserved for more complex cases. Therefore, this design balances the speed and resource efficiency of the persona updating process.

By dynamically capturing and integrating key persona elements, the PCB complements the NSB, thus collectively addressing the dual focus on ``plot'' and ``characters'' in memory extraction.

To accommodate diverse practical deployment needs—such as performance, latency, and memory consumption—our PCB framework is designed to be highly flexible in its choice of language models. It seamlessly supports both general-purpose LLMs and specialized, fine-tuned models, depending on the specific application scenario. As an illustrative example, we demonstrate the use of ChatGPT-4 \cite{gpt4report} to distill the Qwen2-7B model \cite{qwen2.5}, resulting in an optimized model for persona snapshot extraction (details provided in Appendix \ref{app:finetune}). This example highlights the framework’s adaptability, allowing users to leverage either proprietary or open-source LLMs, with or without fine-tuning, based on different requirements.

\subsection{Forgetting Mechanism}
\label{method:forget}

Cognitive science explains forgetting via two main theories: interference (memory traces compete) and inhibitory control (active suppression of irrelevant memories) \cite{lindsay2013humanforgetmechanism1, mensink1988modelforgetmechanism2}. Competition-inhibition mimics human memory by prioritizing relevant information and suppressing outdated or noisy data. For instance, when someone updates their phone number, repeated retrieval of the new number gradually suppresses recall of the old one, leading to its eventual forgetting.

Inspired by this, we model forgetting as a priority-based reallocation, combining temporal decay with inhibitory suppression. When a memory $c$ triggers retrieval from memory pool $M$, we divide memories into:
$\mathbb{R}_c$: relevant and effective memories, 
$\mathbb{N}_c$: interfering (noisy) memories and $\mathbb{U}_c$: unactivated memories.

The mechanism operates from two aspects:

\textbf{Score Computation}. We first define a round $r$ as a pair of consecutive turns in a dialogue, consisting of an utterance and its direct response. Then for every new round, each memory’s importance score $S$ is calculated using 
\begin{equation}
  \label{eq:forget}
  \scalebox{0.95}{$S = \alpha \frac{1}{\exp(\gamma(r_c - b)) + (1 - \epsilon)} + \beta \sum_{r\in R_c} \frac{1}{r_c - r + \epsilon}$},
\end{equation}
where $r_c$ is the current round, $b$ the creation round. $R_c$ is the set of rounds in which the memory is retrieved. $\epsilon$ is a very small value to ensure that the denominator is not 0.
$\alpha$ weighs the first term  temporal decay, reducing the importance of older memories, while $\beta$ weighs the second term models retrieval reinforcement, boosting recently retrieved memories.

\textbf{Retrieval Reinforcement and Suppression}. After calculating all memories scores $S$, we retrieve top $2k$ memories in memory pool. Among them the top $k$ memories are regarded as relevant memories ($\mathbb{R}_c$). We record the current round $r_c$ into the $R_c$ of these memories so that they can be enhanced during the subsequent calculation of $S$. On the other hand, the next $k$ ones (i.e. $\mathbb{N}_c$) will be suppressed by halving their scores $S$, while the unactivated memories (i.e. $\mathbb{U}_c$) remain unchanged. After this step, only memories with higher scores are retained.

This mechanism efficiently prioritizes memory in dialogue systems, preserving relevant content while reducing interference. In practice, we use BGE reranker \cite{bgescore} as our retriever. The parameters $\alpha = 0.1$, $\beta = 0.9$ weigh temporal decay and retrieval reinforcement, respectively. In addition, $k$ is set to 9, which are detailed in Appendix \ref{app:forget_ablation}. 

\section{ZH-4O dataset}

As mentioned earlier, ultra-long conversations play a decisive role in role-playing, but the corresponding datasets are rather scarce. To make up for the deficiency, we construct the ZH-4O dataset, with an average conversation length of 600 turns—significantly exceeding that of existing datasets. Aiming to emulate realistic role-playing scenarios, ZH-4O minimizes restrictions on participants, enabling dialogue annotators and memory labelers to freely exert their creativity and judgment. It is composed of three components: dialogue collection, memory annotation and probe table. 

\subsection{Dialogue Collection}

To construct the ZH-4O dataset, we designed diverse chatbot personas to enhance dialogue variety and generalization. Annotators are briefed on: (1) the chatbot’s persona; (2) adherence to ethics and laws, prohibiting inappropriate content. Each annotator pair completed dialogues of at least 500 turns, with content and style left to their creativity for natural, diverse conversations. Notably, as the dataset is constructed in Chinese, the dialogues reflect distinctive characteristics of the Chinese cultural context. Annotators incorporated culturally specific references, such as Chinese cuisine (e.g., Sichuan cuisine and hotpot) and Chinese celebrities, embedding authentic elements of the Chinese-speaking environment into the conversations. As a result, we collect 28 dialogue sessions, with each session averaging 600 turns in length.


\subsection{Memory Annotation}
After collecting the long dialogues, we assign them to another group of annotators for manual annotation. The annotators are tasked with extracting notable information from the dialogues and indicating the specific turn in which the information appeared. To ensure consistency while minimizing potential bias, we establish a unified workflow for identifying notable information, while allowing flexibility in the specific content and format of the annotations based on annotators' individual judgments. Ultimately, we obtain 1,115 memory annotations. More details on the workflow of annotation are presented in Appendix \ref{app:annotation_workflow}.

\begin{table*}[t]
  \label{mainresults}
  \centering
  \resizebox{0.8\textwidth}{!}{
  \begin{tabular}{lcccccc}
    \hline
    \textbf{Method}& \textbf{BERTScore}& \textbf{M3E}& \textbf{ROUGE-2}& \textbf{ROUGE-L}& \textbf{MemScore} & \textbf{QA precision}\\
    \hline
    Mem0-14B          & 0.6955 & 0.8612 & 0.1542 & 0.4056 & 0.411 & 0.459 \\
    Mem0-32B          & 0.7055 & 0.8672 & 0.1724 & 0.4145 & 0.504 & 0.509 \\
    Mem0-72B          & 0.7108 & 0.8717 & 0.1851 & 0.4288 & 0.519 & 0.612 \\ \midrule
    MemoChat-14B      & 0.7728 & 0.8940 & 0.1973 & 0.3462 & 2.515 & 0.511\\
    MemoChat-32B      & 0.7786 & 0.8884 & 0.2040 & 0.3424 & 2.535 & 0.718\\
    MemoChat-72B      & 0.7788 & 0.8897 & 0.2033 & 0.3404 & 2.303 & 0.693\\ \midrule
    MemoryBank-14B    & 0.7474 & 0.8301 & 0.0292 & 0.0683 & 1.589 & 0.513\\
    MemoryBank-32B    & 0.7591 & 0.8419 & 0.0396 & 0.0830 & 1.732 & 0.651\\
    MemoryBank-72B    & 0.7592 & 0.8374 & 0.0403 & 0.0839 & 1.780 & 0.692\\ \midrule
    MOOM-7B finetuned & \underline{0.8018} & \underline{0.9017} & 0.2806 & 0.4834 & \underline{3.170} & \underline{0.832}\\
    MOOM-14B          & 0.7525 & 0.8912 & 0.2816 & 0.4832 & 2.590 & 0.827\\
    MOOM-32B          & 0.7795 & 0.8931 & \underline{0.2914} & \underline{0.4971} & 2.784 & 0.813\\
    MOOM-72B          & \textbf{0.8071} & \textbf{0.9149} & \textbf{0.3341} & \textbf{0.5153} & \textbf{3.317} & \textbf{0.840}\\ \bottomrule
    \hline
  \end{tabular}
  }
  \caption{\label{cap:mainresults}
    Memory extraction accuracy evaluation of Mem0, MemoChat, MemoryBank and our method MOOM. 14B denotes the use of Qwen1.5-14B as the LLM within the framework, while 32B and 72B refer to Qwen2.5-32B and Qwen2.5-72B, respectively. The MOOM-7B finetuned framework employs a fine-tuned Qwen2-7B model specifically for user character profiling, whereas plot summarization is still conducted using the non-fine-tuned Qwen1.5-14B. The MemScore, evaluated using Qwen2.5-72B, quantifies the similarity between the extracted memory and the corresponding labels, with a maximum score of 5 and a minimum score of 0. All metrics presented in the table follow the principle that higher values indicate better performance.\textbf{Bold} indicates the best performance. \underline{Underline} indicates the second best. 
  }
\end{table*}

\subsection{Memory Probing}

To systematically evaluate memory extraction frameworks, we design a probing task within the ZH-4O dataset. Specifically, annotators are instructed to review dialogue contexts and formulate 1068 multiple-choice questions from the robot’s perspective, each targeting a distinct memory point. Each question is associated with four candidate user responses, among which only one accurately reflects the correct memory information, with the remaining three acting as distractors. This structured probing approach enables us to rigorously assess the effectiveness of different memory frameworks in supporting accurate role-playing dialogues.

In addition, the ZH-4O dataset includes a \textit{Probe Table}, a concise evaluation tool designed to adapt to most role-playing scenarios. For more details, please refer to the Appendix \ref{app:probetable}.

\section{Experiments}

In this section, we outline the experimental setup, including experiments settings, metrics,  and a comprehensive analysis of the experimental results. 

\subsection{Experiments Settings}
To enhance the comparison with our proposed method, we evaluate three representative approaches: Mem0, MemoChat~\cite{lu2023memochat}, and MemoryBank~\cite{memorybank}. Mem0 is an open-source memory extraction plugin designed for general-purpose scenarios. MemoChat facilitates consistent performance in LLMs during extended, open-domain conversations. MemoryBank is tailored for human-computer interactions in roleplay settings. We translate the prompts of Mem0 and MemoChat, which lack native Chinese support, into Chinese to ensure a fair and consistent evaluation.

In the memory extraction experiments, both our proposed MOOM framework and the baseline methods rely on LLMs for memory extraction. To investigate the performance of different frameworks across LLMs with varying parameter scales, we utilize Qwen1.5-14B \cite{qwenreport}, Qwen2.5-32B \cite{qwen2.5}, and Qwen2.5-72B as the primary LLMs in our experiments. Additionally, as discussed earlier, we fine-tune the Qwen2-7B model based on results from GPT-4, making it a specialized model for character profile extraction within the MOOM framework. All experiments are repeated three times and the average results are reported.
The experiments were conducted on a eight NVIDIA H800 GPUs to accommodate the computational requirements of a 72B parameter model. For experiments involving models with smaller parameter sizes, such as 7B, execution was typically performed on a single NVIDIA H800 GPU.

\subsection{Metrics}

To evaluate memory extraction accuracy, let $M$ be the set of extracted memories, $L$ the human-annotated memory labels, and $U$ all dialogue information. As $L$ reflects human preferences, we prioritize a scoring function $f(l,M)$ to measure how well $M$ aligns with $L$, over $g(m,L)$, which focuses on the relevance of memories $m$ to labels, better capturing human-centric priorities.

We employ multiple evaluation metrics to quantify the accuracy of memory extraction, including \textbf{BERTScore} \cite{bertscore} precision, \textbf{M3E} \cite{m3e} embedding similarity, and \textbf{ROUGE} \cite{lin2004rouge} precision. For each label $l\in L$, we calculate its similarity score with all memories in $M$ using these metrics and select the highest similarity score as the score for that label. Finally, we average the scores $f$ of all labels to obtain the final score for the framework.


To evaluate memory extraction fidelity, we introduce \textbf{MemScore}, a metric based on large language model (LLM) assessments. For each memory $m \in M$, we compute its BERTScore similarity with reference labels $L$, selecting the highest-scoring label as the best match. An LLM then scores (0–5) how well $m$ captures the label’s semantic intent, retaining the highest score for repeated labels. Unmatched labels receive a score of 0 to penalize incomplete recall, ensuring scalable and robust evaluation.

Besides, we employ probe questions from the ZH-4O dataset. After providing the extracted information, we prompt the model to determine the optimal choice for each probe question. The model’s selections are then compared against the labeled ground-truth answers to calculate the precision metric. This precision score serves as an indicator for assessing the performance of memory plugins.

\begin{figure}[t]
    \centering
    \begin{subfigure}[b]{0.23\textwidth}
        \centering
        \includegraphics[width=\textwidth]{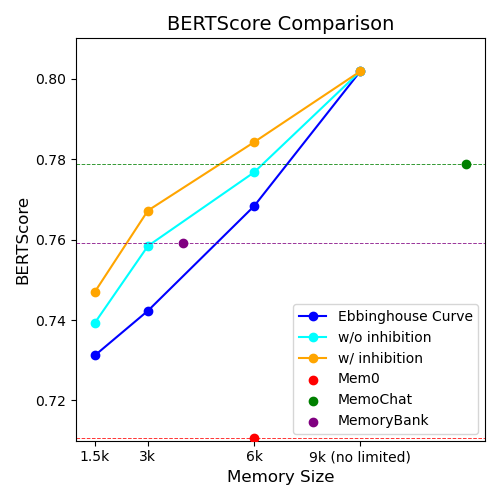} 
        \label{fig:bertscore}
    \end{subfigure}
    \hfill 
    \begin{subfigure}[b]{0.23\textwidth}
        \centering
        \includegraphics[width=\textwidth]{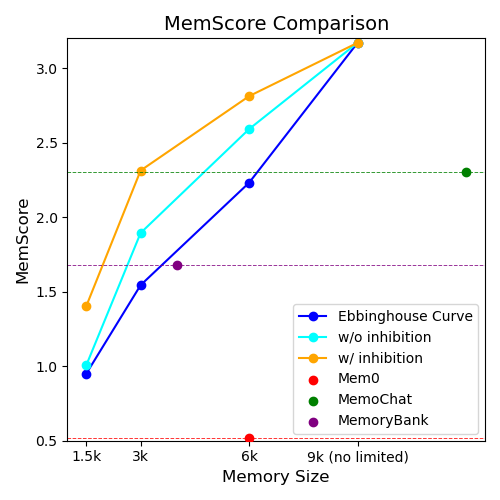} 
        \label{fig:qaScore}
    \end{subfigure}

    \caption{Results of our forget mechanism experiments. Ebbinghause curve, without inhibition strategy method and ours' competition-inhibition method (w/ inhibition) are in the finetuned MOOM-7B framework. Other methods use Qwen2.5-72B.} 
    \label{fig:forgetexp}
\end{figure}

\subsection{Memory Extraction Accuracy Evaluation}
\label{sec:mem_extract_experiment}
The results in Table \ref{cap:mainresults} demonstrate that the MOOM framework consistently exhibits superior memory extraction accuracy under different model capacities. Notably, with the fine-tuned 7B model, MOOM achieves significantly higher scores compared with other frameworks, and also obtains a relatively high LLM score (3.170). This indicates that adopting a task-specific fine-tuned model within the MOOM framework effectively enhances the extraction of critical memory information.



For larger models (e.g., 32B), MOOM significantly boosts LLM scores (2.784) and outperforms in ROUGE-L and M3E, demonstrating its robustness across scales. This is due to its detailed key-value summaries, which enhance large models' summarization capabilities. Even with an untuned small model (14B), MOOM achieves comparable or superior performance to frameworks using much larger models (72B), highlighting its effectiveness in memory extraction.

In terms of time efficiency, a H800 GPU process 20 rounds of dialogue requires approximately 8 seconds for 7B model. 
When utilizing the same LLM, the time consumption ratio of MOOM compared with Mem0, MemoChat, and MemoryBank is approximately 1:2.5:3:1.5, respectively. This demonstrates that MOOM not only maintains high accuracy but also significantly reduces memory extraction latency.

MOOM also outperforms baselines on the English LoCoMo dataset, demonstrating its applicability beyond Chinese dialogues. The detailed results can be found in the Appendix \ref{app:locomoexperiment}.

\subsection{Forget Mechanism Evaluation}

In Fig. \ref{fig:forgetexp}, it can be observed that within the MOOM framework, both the Ebbinghaus forgetting strategy and the proposed competition-inhibition strategy outperform other methods under comparable memory capacities. For example, in terms of BERTScore, MOOM surpasses MemoryBank at a memory size of 3k and outperforms Mem0 at 6k. Notably, our competition-inhibition forgetting algorithm consistently achieves better performance than the Ebbinghaus strategy. More results are presented in Appendix \ref{app:supp_forget}. 

These results demonstrate that the proposed competition-inhibition forgetting algorithm enables flexible adjustment of the memory capacity within the MOOM framework. Compared to other forgetting strategies, our method attains a more efficient balance between memory retention and forgetting. This makes it a superior solution for memory management in ultra-long dialogue scenarios, ensuring both adaptability and effectiveness. 

\begin{table*}[t]
  \label{tab:ablation}
  \centering
  \resizebox{0.8\textwidth}{!}{
  \begin{tabular}{lcccccc}
    \hline
    & \textbf{BERTScore}& \textbf{M3E}& \textbf{ROUGE-2}& \textbf{ROUGE-L} &\textbf{MemScore}&\textbf{QA precision}\\
    \hline
    NSB Only       & 0.7729 & 0.8569 & 0.0535 & 0.1058 & 2.603 & 0.693 \\
    PCB Only         & 0.7898 & 0.9021 & 0.2376 & 0.4224 & 2.468 & 0.752 \\
    MOOM                 & 0.8018 &	0.9017 & 0.2806 & 0.4834 & 3.170 & 0.832 \\
    \hline
  \end{tabular}
  }
  \caption{\label{cap:ablation}
    Results of the ablation study for the MOOM framework.
  }
\end{table*}

\begin{table}[h]
  \centering
  \resizebox{0.4\textwidth}{!}{
  \begin{tabular}{lccccc}
    \hline
    & \textbf{vanilla} & \textbf{InfLLM} & \textbf{RAPTOR} & \textbf{HippoRAG2} & \textbf{MOOM}\\
    \hline
    Qwen1.5-7B & 0.607 & 0.570 & 0.626 & 0.730  & \textbf{0.793}\\
    Qwen1.5-14B & 0.687 & 0.635 & 0.717 & 0.763 & \textbf{0.827}\\
    Qwen2.5-1M-7B & 0.781 & 0.673 & 0.722 & 0.759 &  \textbf{0.831}\\
    Qwen2.5-1M-14B & \textbf{0.852} & 0.703 & 0.740 & 0.792 & 0.836\\
    \hline
  \end{tabular}
  }
  \caption{\label{cap:qaprobe}
    Probe-based QA evaluation: All the model use Qwen1.5-14B as base model. The context window of vanilla Qwen1.5, InfLLM, RAPTOR, HippoRAG2 and MOOM are set to 8k, 2k and 1k. The context window of vanilla Qwen2.5-1M is no limited.
  }
\end{table}

\subsection{Comparison with Long Context Methods}
To assess the efficacy of memory retrieval in enhancing dialogue model performance, we compare our proposed MOOM with InfLLM \cite{xiao2024infllm}, a long-context model employing sliding-window attention and context memory modules, and the vanilla Qwen model. The vanilla Qwen model concatenates dialogues and questions into a single prompt, as does InfLLM within its framework. In contrast, MOOM integrates only the recalled memory with the questions, excluding the original dialogues. Specifically, MOOM’s context window incorporates nine retrieved memory contents.

As presented in Table \ref{cap:qaprobe}, MOOM achieves superior performance compared to both InfLLM and vanilla Qwen, despite utilizing significantly shorter context window. For instance, the vanilla Qwen1.5-7B model, with an 8k context window, requires approximately 32GB of GPU memory. InfLLM lowers the GPU memory demand to around 16GB. MOOM, accounting for the memory required during retrieval, also operates efficiently with approximately 16GB memory. This demonstrates that MOOM is well suited for resource-constrained environments with limited GPU capacity.

For further comparison, we evaluate MOOM against newer models designed for ultra-long contexts, such as Qwen2.5-7B-1M and Qwen2.5-14B-1M, which support unrestricted context windows (e.g., up to 32k tokens for the longest dialogues in ZH-4O). These models achieve QA probing scores of 0.781 and 0.852, respectively. Despite the extended context capabilities of these models, MOOM remains highly competitive. Notably, in practical applications, MOOM’s memory retrieval framework can operate in parallel with the dialogue model, significantly reducing memory overhead. In contrast, directly processing entire dialogues with ultra-long-context models demands substantial GPU memory, making MOOM a more resource-efficient solution for open-domain dialogue systems.

\subsection{Ablation study}

To validate the effectiveness of MOOM's dual-branch design, we conduct ablation experiments by isolating the two branches: NSB and PCB. The results are summarized in Table \ref{cap:ablation}.

The complete MOOM framework, integrating both branches, achieves the highest performance across all metrics. When evaluating the branches independently, distinct strengths become evident. The narrative summarization-only branch demonstrates superior performance in MemScore (2.603), outperforming the persona-only branch (2.468). This advantage suggests that NSB effectively captures interconnected memory points reflective of coherent plot narratives, closely aligning with memory structures evaluated by MemScore.

Conversely, the persona-only branch exhibits stronger performance in QA precision (0.752), surpassing the narrative-only branch (0.693). This result likely arises because PCB separately encodes individual informational elements from dialogues, which better aligns with the probe-style evaluation employed by QA precision.

These results highlight the complementary strengths of the NSB and PCB branches within MOOM. 
More ablation studies are represented in Appendix \ref{app:pcb_ablation}.

\section{Conclusion}

In this paper, we propose MOOM, a dual-branch memory extraction framework for role-playing in ultra-long dialogues. To mitigate uncontrolled memory growth, we propose a forgetting mechanism based on ``competition-inhibition'' theory. Additionally, we present ZH-4O, a Chinese ultra-long dialogue dataset with high-quality annotated memory. Experimental results show that MOOM exceeds the SOTA methods in accuracy, efficiency, and memory control.

\section{Limitations}
\textbf{Annotation Diversity.}  One limitation of our study lies in the diversity of the annotation team. Our annotation team consists of 10 members, 6 females and 4 males. Since the ZH-4O dataset is in Chinese, all annotators are Chinese. In addition, all annotators received a high level of education, which may not fully represent the general population. Although ZH-4O is designed to include various types of intimate relationships, privacy information such as personal backgrounds do not be collected. This lack of demographic diversity may introduce bias, potentially limiting the dataset's generalizability to real-world scenarios.

\textbf{Language.}  Our research primarily focuses on the Chinese. While the proposed MOOM framework demonstrates strong performance on both Chinese (ZH-4O) and English (LoCoMo) datasets, our ZH-4O dataset is exclusively in Chinese, which limits its linguistic diversity. In future work, we plan to expand the ZH-4O dataset to include more languages, enhancing its applicability to multilingual dialogue scenarios.

\textbf{Practical Applications.}  Through fine-tuning, our approach enables a 7B-parameter model to outperform larger models in other frameworks, and the overall architecture supports asynchronous invocation. However, we finetune the 7B-parameter model based on GPT-4 output data only, potentially introducing biases specific to GPT-4's outputs. Consequently, it is essential for practitioners employing our framework to acknowledge this limitation. When adapting smaller-scale models, careful consideration should be given to dataset construction. Depending on the application's requirements and desired outcomes, practitioners may select alternative data sources beyond GPT-4's generated outputs, such as utilizing large-scale open-source models or manually curated datasets. Diversifying training data sources in this manner can mitigate the risk of embedding specific biases inherent to any single model or data generation method.

\textbf{Dialogue and Role-Playing.}
The use of human-AI dialogue is a prevalent approach for leveraging LLMs in role-playing scenarios, but it represents only one of many possible modalities. The ZH-4O dataset, as constructed, is limited to text-based, two-party human-AI dialogues, which restricts its scope. It is possible to consider integrating multimodal elements, such as images or audio, or support multi-party role-playing scenarios involving multiple characters. 

In future work, we aim to address these constraints by exploring more flexible and diverse role-playing frameworks, incorporating multimodal inputs and multi-agent interactions to enhance the immersive quality and applicability of the ZH-4O dataset.

\bibliography{acl_latex}

\appendix

\section{Annotation Workflow of the ZH-4O Dataset}
\label{app:annotation_workflow}
The ZH-4O dataset is constructed through a meticulous annotation process designed to ensure the quality and diversity of the collected dialogue data. This chapter outlines the comprehensive workflow, which encompasses dialogue generation, memory annotation, and probe annotation. The annotation team consists of 10 members, with 6 female and 4 male annotators.

\subsection{Dialogue Generation}

The dialogue generation phase leverages a specialized model tailored for role-playing conversations. A subset of the annotation team, comprising 3 female and 2 male annotators, is selected to participate in this phase. Prior to the task, the annotators are briefed on the project's academic purpose and its eventual public release. Clear instructions are provided to safeguard personal privacy and prohibit the inclusion of inappropriate content, such as explicit, violent, or hateful material. Annotators are informed that they can withdraw from the task at any time. To preserve the naturalness and spontaneity of the dialogues, minimal constraints are imposed on the conversation content, allowing annotators to engage freely with the LLM agent.

Upon consenting to participate, annotators are provided with basic information about the role assumed by the dialogue model. They then engage in open-ended conversations, which are recorded in real-time in the backend system. This setup ensures the capture of diverse and authentic conversational data, forming the foundation of the ZH-4O dataset.

\subsection{Memory Annotation}

Following the dialogue generation phase, the remaining 3 female and 2 male annotators are tasked with memory annotation. This process aims to extract and tag significant information from the generated dialogues. The workflow for memory annotation is structured as follows.

\begin{enumerate}
    \item \textbf{Initial Review}: Annotators rapidly skim the entire dialogue to gain an overview of its content and context.
    \item \textbf{Detailed Extraction}: Annotators then review the dialogue sentence by sentence, identifying and extracting information deemed important. Each piece of extracted information is tagged with its corresponding position in the dialogue.
    \item \textbf{Dual Annotation and Review}: Each dialogue is independently annotated by two annotators. A third annotator subsequently reviews the annotations to identify discrepancies, correct errors, and integrate memory points with similar meanings.
\end{enumerate}

This unconstrained annotation approach gives the ZH-4O dataset greater subjectivity and diversity in its memory annotations. We observe that the majority of the memory annotations follow a subject-verb-object (SVO) structure, such as ``Weifu enjoys observing the night sky''. However, it should be noted that it is not encouraged to artificially fit the SVO structure of the annotated memory information for the sake of achieving higher evaluation scores. Instead, it is more worth focusing on how to utilize these memory annotations in meaningful ways.

\begin{figure*}[ht]
    \includegraphics[width=\textwidth]{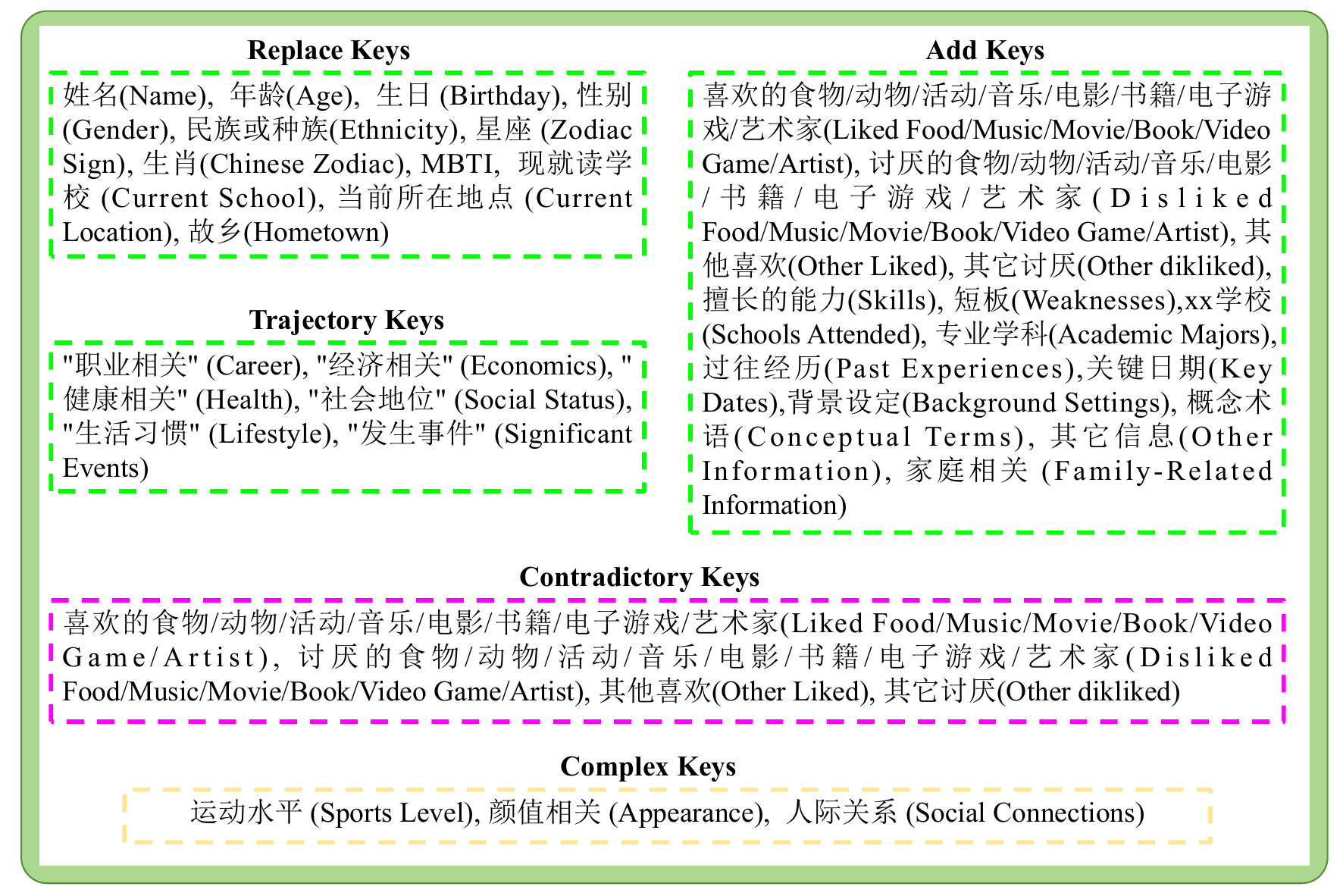}
    \caption{Illustration of the Persona Branch’s merging mechanism. }
    \label{fig:appendix_kv}
\end{figure*}

\begin{figure}[h]
    \centering
    \includegraphics[width=0.5\textwidth]{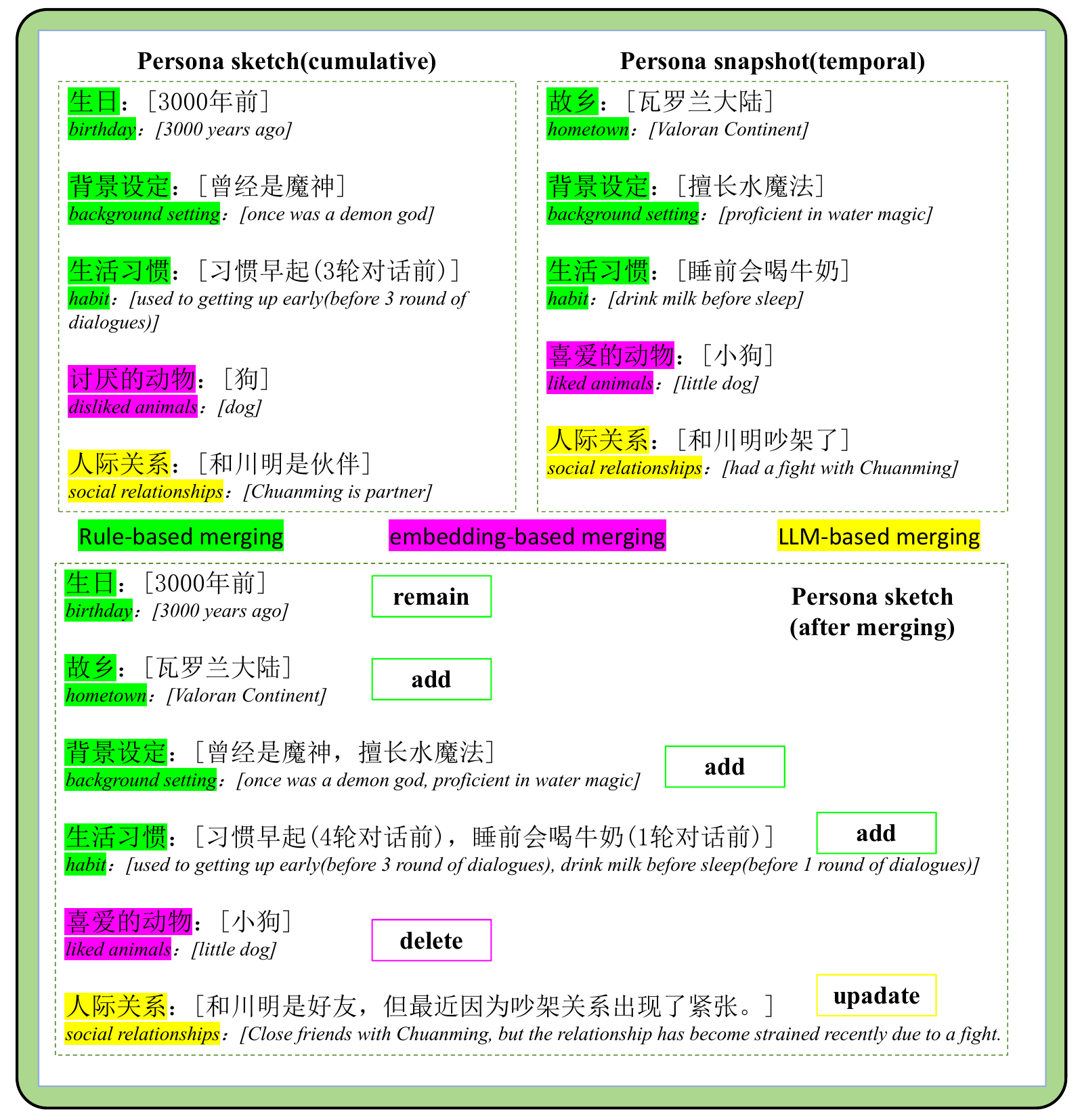}
    \caption{A case of the Persona Branch’s merging mechanism. The persona information is divided into two categories: a history cumulative persona sketch (up left) and a temporal persona snapshot (up right). The merging process integrates information from both sources using three methods. The final persona sketch after merging (bottom) ensures a coherent and up-to-date character profile.}
    \label{fig:case_kv}
\end{figure}

\subsection{Probe Annotation}

The probe annotation phase consists of two distinct components: regular probe annotation and universal probe annotation. These tasks are designed to create question-answer pairs and probe tables to evaluate the dialogue model's comprehension and robustness.

\subsubsection{Regular Probe Annotation}

In the regular probe annotation task, a designated reviewer utilizes the finalized memory information table from the memory annotation phase. The reviewer selects key memory points and designed multiple-choice question-answer (QA) pairs, each with four options. Only one option is correct, while the remaining three serve as distractors.

\subsubsection{Universal Probe Annotation}

The universal probe annotation involves the collaborative efforts of 3 annotators to construct a comprehensive probe table. The probe table consists of multiple probe items, each comprising:

\begin{itemize}
    \item \textbf{Probe Information}: A specific piece of information inserted into the initial dialogue.
    \item \textbf{Probe Question}: A question posed at a later point in the dialogue to test the model's retention of the probe information.
    \item \textbf{Reference Answer}: The expected response to the probe question for comparison with the model output.
\end{itemize}

To ensure the applicability of the probe table, it is designed to cover a wide range of everyday scenarios. A probe item is implemented by inserting the probe information into the dialogue and posing the corresponding question. The model response is then compared to the reference answer to assess its performance.

After the probe table is completed, an additional review is conducted by two other annotators. This review verified the content of the probe table and ensured that the probe questions are mutually independent, preventing overlap or interference between items.

\section{Persona Keys Definitions and Merging Strategies}
\label{app:personakeys}
To construct and update personas, our framework organizes key-value pairs into four categories, each with specific merging strategies. All the keys are in Fig \ref{fig:appendix_kv}.

\begin{itemize}
    \item[(1)]
    \textbf{Replace Keys}: Rule based merging. These keys are associated with static attributes that are unlikely to change or be expanded over time. When a new value is extracted for these keys, the old value is replaced directly. 
    \item[(2)]
    \textbf{Information-retrieval dialogue $Q$}: These keys represent expandable attributes, such as preferences, skills, or background information. New values are appended to the existing list of values without modification.
    \item[(3)]
    \textbf{Trajectory Keys }: Rule-based merging. Trajectory keys are updated by dynamically adjusting timestamps to reflect the recency of each value relative to the current dialogue turn. When new values are introduced, they are appended to the existing list with the current timestamp, while the timestamps of older values are incremented to reflect their increasing distance from the present.
    \item[(4)]
    \textbf{Contradictory Keys}: Embedding-based merging. See Section \ref{method:kv} for more details.
    \item[(5)]
    \textbf{Complex Keys}: LLM-based merging. See Section \ref{method:kv} for more details.
\end{itemize}

The merging case is shown in Figure \ref{fig:case_kv}.

\begin{figure*}[h]
    \centering
    \begin{subfigure}[b]{0.3\textwidth}
        \centering
        \includegraphics[width=\textwidth]{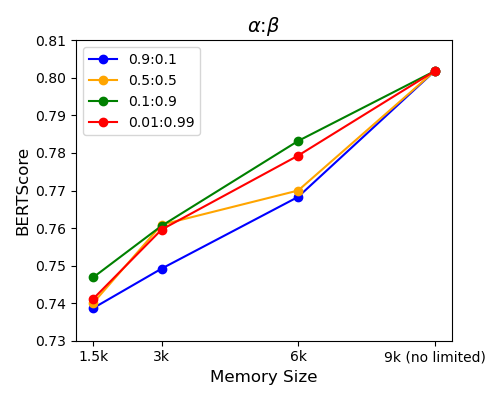} 
        \label{fig:alpha_beta_ablation}
    \end{subfigure}
    \hfill 
    \begin{subfigure}[b]{0.3\textwidth}
        \centering
        \includegraphics[width=\textwidth]{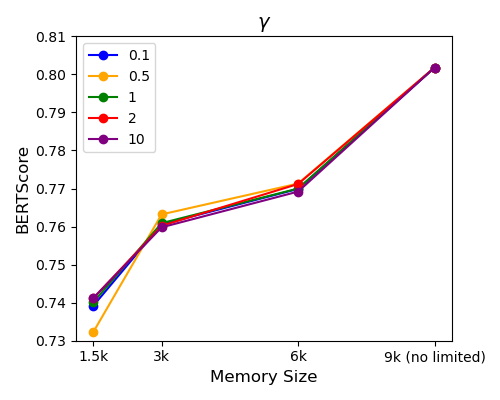} 
        \label{fig:gamma_ablation}
    \end{subfigure}
    \hfill
    \begin{subfigure}[b]{0.3\textwidth}
        \centering
        \includegraphics[width=\textwidth]{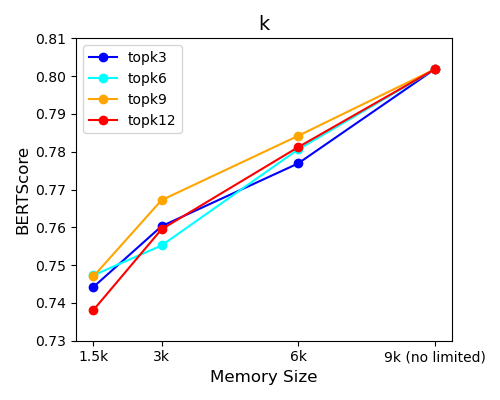} 
        \label{fig:k_ablation}
    \end{subfigure}
    \caption{Results of our forget mechanism parameters experiments. All methods are in the finetuned MOOM-7B framework. The up left figure illustrates how different proportions of the $\alpha$ and $\beta$ parameters, constrained to sum to 1, affect the BERTScore performance. The up right figure examines the influence of varying $\gamma$ values on BERTScore, providing insights into the sensitivity of the model to this parameter. The below presents how $k$ influence performance of our forget mechanism.} 
    \label{fig:forgetablation}
\end{figure*}

\begin{figure*}[t]
    \centering
    \begin{subfigure}[b]{0.4\textwidth}
        \centering
        \includegraphics[width=\textwidth]{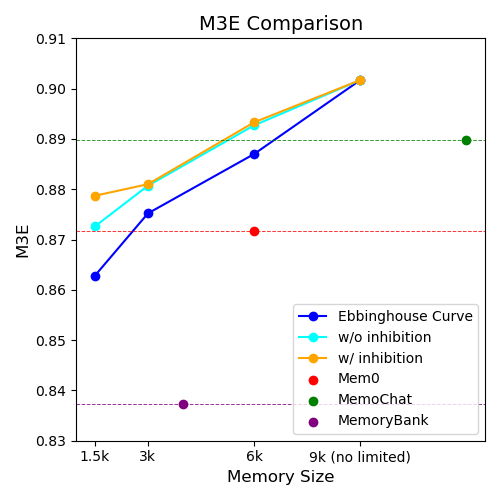} 
        \label{fig:m3e}
    \end{subfigure}
    \hfill 
    \begin{subfigure}[b]{0.4\textwidth}
        \centering
        \includegraphics[width=\textwidth]{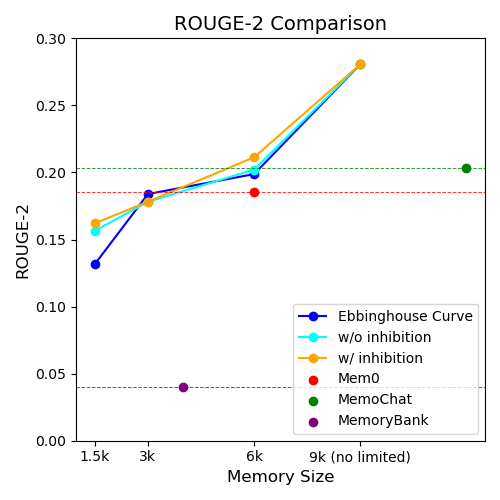} 
        \label{fig:rougeL}
    \end{subfigure}
    \hfill 
    \begin{subfigure}[b]{0.4\textwidth}
        \centering
        \includegraphics[width=\textwidth]{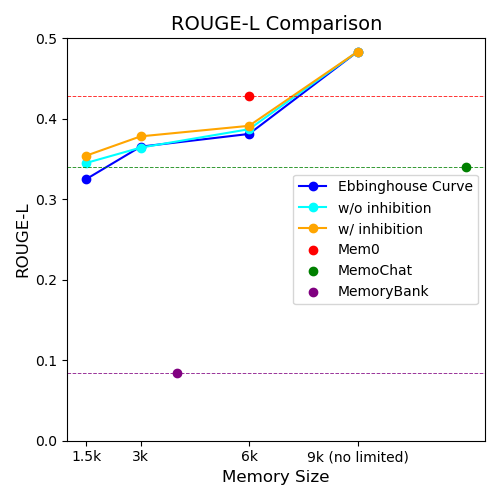} 
        \label{fig:rougeL}
    \end{subfigure}
    \hfill 
    \begin{subfigure}[b]{0.4\textwidth}
        \centering
        \includegraphics[width=\textwidth]{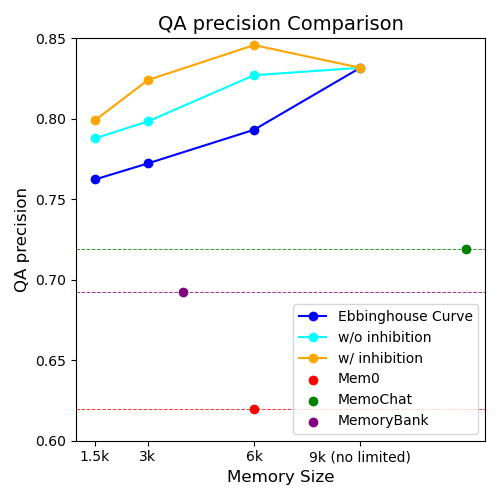} 
        \label{fig:rougeL}
    \end{subfigure}

    \caption{Results of our forget mechanism experiments on M3E, ROUGE-L, ROUGE-2 and QA precision. Ebbinghaus curve and ours ``competition-inhibition'' method are in the finetuned MOOM-7B framework. Other methods use Qwen2.5-72B.} 
    \label{fig:forgetsupplement}
\end{figure*}

\section{Persona Extraction Model Finetuning}
\label{app:finetune}
In this section, we briefly introduce how we finetune a Qwen2-7B model to extract a persona snapshot.
    \begin{itemize}
    \item[(1)]
    \textbf{Data Collection and Preprocessing}: We invite approximately 100 experienced users to engage in human-machine dialogues. These dialogues are segmented into 10-turn slices to ensure manageable and contextually relevant chunks of data. This preprocessing step yields an initial pool of 10,000 dialogue slices.
    \item[(2)]
    \textbf{Persona Annotation with GPT-4}: Each dialogue slice is processed using GPT-4 to generate summarized persona snapshots. The generated outputs are then carefully reviewed and refined to ensure accuracy, coherence, and consistency. This cleaning process eliminates noise and ensures that the labels are of high quality, suitable for supervised fine-tuning.
    \item[(3)]
    \textbf{Fine-tuning Configuration}: The cleaned persona summaries are used as labels for fine-tuning the Qwen2-7B model. The training process employs a batch size of 64, a learning rate of 1e-05, and spans 800 steps. This configuration is chosen to balance training efficiency with model performance, ensuring convergence without overfitting.
    \end{itemize}

\section{Ablation Study on Forgetting Mechanism Parameters}
\label{app:forget_ablation}

We further explored the impact of parameter selection in the forgetting mechanism on memory extraction performance. As shown in Fig. \ref{fig:forgetablation}, the optimal values for $\alpha$ and $\beta$ are 0.1 and 0.9, respectively. Notably, the best-performing $\alpha$ and $\beta$ values are not balanced, suggesting that retrieval-induced forgetting should be given greater consideration than time-induced forgetting in our framework. However, assigning excessive weight to retrieval-induced forgetting may lead to a decline in overall algorithm performance. While the choice of $\alpha$ and $\beta$ significantly affects the forgetting mechanism, the $\gamma$ parameter exhibits minimal influence within a limited range. Consequently, we set $\gamma$ to 1 as a balanced choice.

In addition, as presented in Fig. \ref{fig:forgetablation}, the optimal value for $k$ is 9. This finding aligns with our observation, where the accuracy of QA responses increased with the number of recalled memories when fewer than nine memories are retrieved. Beyond nine memories, further increases in the number of recalled memories yielded no significant improvement in QA performance. This observation is consistent with the ``inhibition-competition'' theory, suggesting that $k$ likely represents the threshold for relevant information critical to effective memory retrieval.

\section{Supplement to Forget Mechanism Evaluation}
\label{app:supp_forget}

Fig \ref{fig:forgetsupplement} shows the results of the forget mechanism in M3E and ROUGE-L. These two metrics are both higher-the-better.
Our method outperforms both the forgetting algorithm based on the Ebbinghaus curve and other algorithms of the same memory capacity across both evaluation metrics. Notably, it maintains superior performance even with a lower memory capacity. For example, at a capacity of 6k, our method achieves higher scores on both M3E and ROUGE-L compared to MemoChat with a capacity of 10k, which aligns with our conclusions in Fig. \ref{fig:forgetexp}. However, we also observe that at capacity of 6k, our method is worse than  Mem0 in terms of ROUGE-L. We hypothesize that this discrepancy arises because Mem0 employs a memory representation that more closely resembles a subject-verb-object structured memory tagging system. Furthermore, it is noteworthy that using the competition-inhibition strategy, a limited memory capacity (6k) achieves higher QA precision than an unlimited memory setting. We attribute this to the inhibition-competition mechanism's ability to effectively suppress noisy information, thereby enhancing the relevance and utility of the retrieved content.

\begin{table*}[t]
  \label{tab:pcb_ablation}
  \centering
  
  \begin{tabular}{lccccc}
    \hline
    & \textbf{BERTScore}& \textbf{M3E}& \textbf{ROUGE-2}& \textbf{ROUGE-L} &\textbf{MemScore}\\
    \hline
    Rule-Based Only  & 0.7989 &	0.8992 & 0.2701 & 0.4694 & 2.843  \\
    +Embedding-Based  & 0.8015 & 0.9004 & 0.2787 & 0.4782 & 3.172 \\
    +LLM-Based       & 0.8018 &	0.9017 & 0.2806 & 0.4834 & 3.170 \\
    \midrule
    Rule-Based Only\textsuperscript{\textdagger}  & 0.7107 &	0.8747 & 0.1301 & 0.3342 & 1.189 \\
    +Embedding-Based\textsuperscript{\textdagger}  & 0.7214 & 0.8745 & 0.1325 & 0.3491 & 1.205  \\
    +LLM-Based\textsuperscript{\textdagger}       & 0.7470 &	0.8787 & 0.1621 & 0.3543 & 1.405 \\
    \hline
  \end{tabular}
  
  \caption{\label{cap:pcb_ablation}
    The ablation study on the persona construction branch. \textsuperscript{\textdagger} denotes the variant with limited memory capacity, where the number of stored memory items is constrained to 50.
  }
\end{table*}
\section{Ablation Study on Persona Construction Branch}
\label{app:pcb_ablation}
Table \ref{cap:pcb_ablation} presents the results of the ablation study on the Persona Construction Branch (PCB). When memory capacity is unconstrained, incorporating Embedding-Based and LLM-Based merging strategies on top of Rule-Based merging yields marginal improvements in memory retrieval performance across all metrics. Specifically, the addition of Embedding-Based merging increases MemScore from 2.843 to 3.172, while LLM-Based merging achieves a comparable MemScore of 3.170.

Under constrained memory conditions, where the number of stored memory items is limited to 50, both Embedding-Based and LLM-Based merging strategies demonstrate more substantial performance gains compared to the Rule-Based Only baseline. Notably, LLM-Based merging achieves the highest improvement, increasing MemScore from 1.189 to 1.405 and ROUGE-2 from 0.1301 to 0.1621. We hypothesize that LLM-Based merging excels at consolidating multiple pieces of information into a single, coherent memory item. While this capability offers limited advantages in unconstrained memory settings, it becomes critical for effective memory extraction and retention when memory capacity is restricted.

\begin{table*}[h]
  \label{locomoresults}
  \centering
  
  \begin{tabular}{lccccc}
    \hline
    \textbf{Method}& \textbf{BERTscore}& \textbf{ROUGE-2}& \textbf{ROUGE-L}& \textbf{MemScore}\\
    \hline
    Mem0-14B          & 0.5932 & 0.1232 & 0.3856 & 0.458 \\
    Mem0-72B          & 0.6312 & 0.1442 & 0.3904 & 0.524 \\ \midrule
    MemoChat-14B      & 0.6223 & 0.1563 & 0.3061 & 2.362 \\
    MemoChat-72B      & 0.6872 & 0.1733 & 0.3198 & 2.521 \\ \midrule
    MemoryBank-14B    & 0.6247 & 0.0387 & 0.0684 & 1.490 \\
    MemoryBank-72B    & 0.6712 & 0.0433 & 0.0981 & 1.680 \\ \midrule
    MOOM-14B          & 0.6521 & 0.2492 & 0.4211 & 2.324 \\
    MOOM-72B          & \textbf{0.7088} & \textbf{0.2925} & \textbf{0.4288} & \textbf{3.288} \\ \bottomrule
    \hline
  \end{tabular}
  
  \caption{\label{cap:locomoresults}
    Memory extraction accuracy evaluation of Mem0, MemoChat, MemoryBank and our method MOOM on LoCoMo dataset. 14B denotes the use of Qwen1.5-14B as the LLM within the framework, while 72B refer to Qwen2.5-72B, respectively. The MemScore is evaluated using Qwen2.5-72B. All metrics presented in the table follow the principle that higher values indicate better performance.
  }
\end{table*}

\section{Experiments on LoCoMo}
\label{app:locomoexperiment}

We also conduct experiments on the LoCoMo dataset. The basic setup is similar to that described in Section \ref{sec:mem_extract_experiment}. However, we made the following adjustments specific to the LoCoMo dataset:
First, the LoCoMo dataset segments long dialogues into sessions of varying lengths. In our experiments, we treat each session as a unit when extracting memories.
Second, the LoCoMo dataset provides three types of labels: event summary, observation, and session summary. Given the considerable length of the session summary, we utilize only the event summary and observation as memory labels.
Third, since LoCoMo is an English multimodal long-dialogue dataset, we standardize our prompts by translating them into English (or using the original English prompts when available). In addition, we ignore any images present in the dialogues during the testing.

As shown in Table \ref{cap:locomoresults}, even in the English LoCoMo dataset, our approach, MOOM, outperforms other methods. This demonstrates the cross-linguistic robustness of our model. In particular, MOOM achieves superior metrics compared to competing methods, regardless of whether it is built on the 14B model or the 72B model.

\section{Probe Table}
\label{app:probetable}

To efficiently evaluate the impact of memory extraction frameworks on model response quality—while avoiding the high cost of manual QA annotations—we design a \textbf{probe table} as a standardized evaluation tool. This table includes 30 structured sets, each consisting of:

\begin{itemize}
    \item \textbf{Information-containing dialogue ($P$):} a user utterance implicitly embedding the target information.
    \item \textbf{Information-retrieval dialogue ($Q$):} a follow-up dialogue designed to trigger memory recall.
    \item \textbf{Reference answer ($A$):} the ideal model response assuming successful memory retrieval from $P$.
\end{itemize}

In evaluation, $P$ and $Q$ are inserted into the dialogue flow. The model’s response to $Q$ is then compared with $A$ to assess how well the framework supports accurate recall.

The probe sets are carefully curated to be mutually independent, ensuring that each $(P, Q, A)$ triplet conveys a distinct piece of information without semantic overlap or interference with others. For instance, if one probe concerns a user's favorite fruit, no other probe will reference preferences or content related to food, thereby minimizing cross-probe redundancy and contamination.

\begin{table}[t]
  \centering
  \resizebox{0.45\textwidth}{!}{
  \begin{tabular}{cccc}
    \hline
    \textbf{MOOM} & \textbf{Mem0} & \textbf{MemoChat} & \textbf{MemoryBank} \\
    \hline
    1.521 & 0.951 & 1.212 & 0.942 \\
    \hline
  \end{tabular}
  }
  \caption{\label{cap:probe}
    Probe-based QA evaluation: similarity scores between generated responses and ground-truth answers (scale 0–5). MOOM uses fine-tuned Qwen2-7B; others use Qwen2.5-72B.
  }
\end{table}

To assess the practical utility of extracted memories in enhancing response generation, we adopt a QA-style evaluation framework built upon the $(P, Q, A)$ triplets from the probe table. Given a retrieval prompt $Q$, relevant memories are selected using BGE and combined with limited dialogue context to guide Qwen2.5-72B in generating a response. A second Qwen2.5-72B agent then evaluates the semantic similarity between the generated response and the reference answer $A$, scoring it on a 0–5 scale. Notably, while all frameworks employ Qwen2.5-72B for this process, the MOOM system utilizes a fine-tuned Qwen1.5-7B.

As summarized in Table~\ref{cap:probe}, MOOM achieves the highest QA evaluation score, indicating its superior capability in extracting and utilizing relevant memory content. Although MemoryBank ranks highly in intrinsic memory accuracy (BERTScore and MemScore), it performs the worst in this QA-based setting, likely due to its tendency to include excessive redundant information (see Appendix~\ref{app:casestudy}). In contrast, MemoChat, which shares a key-value structure with MOOM, attains the second-best performance, suggesting that key-value memory representations may offer a more effective paradigm for dialogue-centric memory retrieval.

\section{Case Study}
\label{app:casestudy}

\begin{figure*}[h]
    \includegraphics[width=\textwidth]{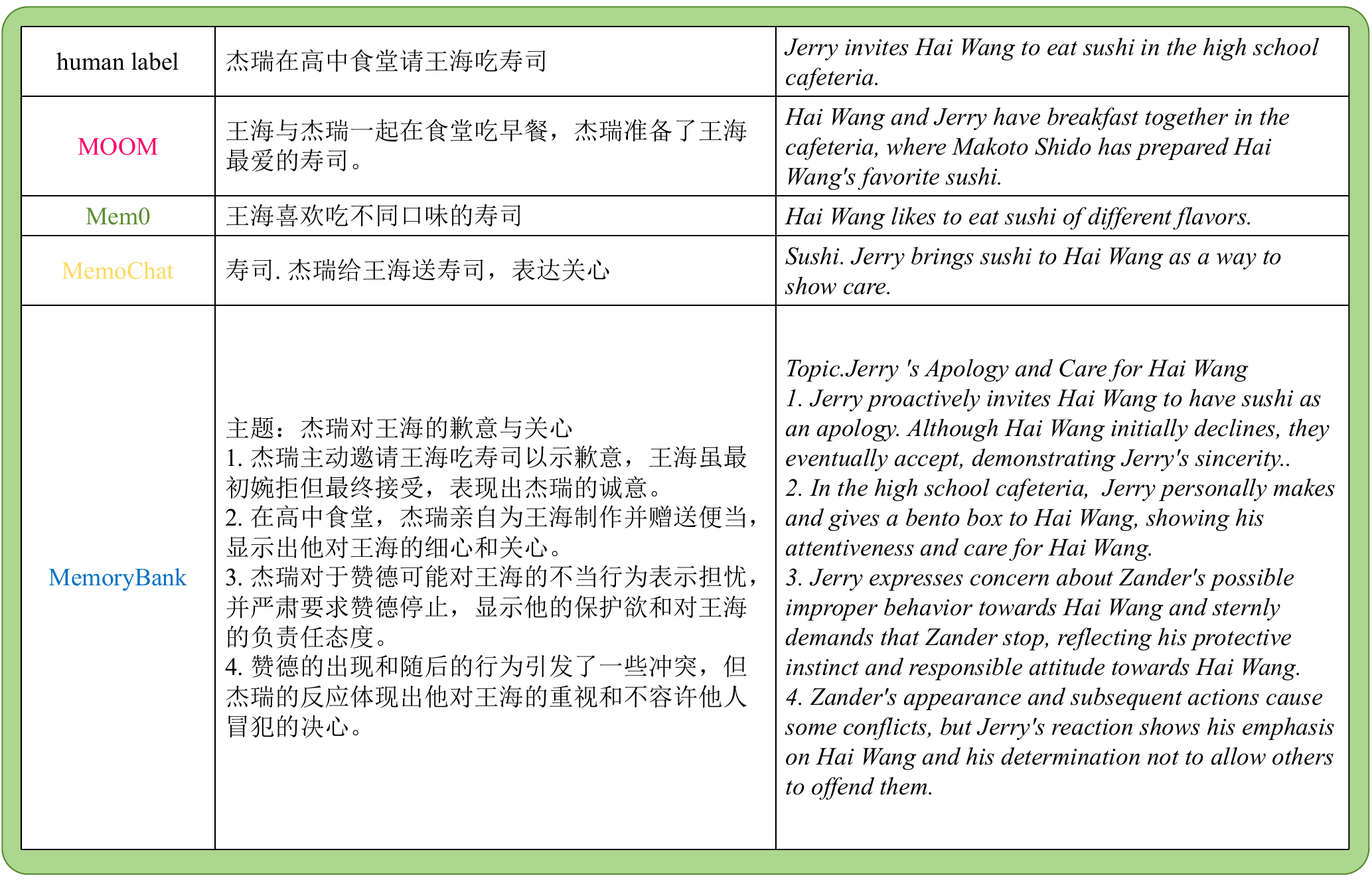}
    \caption{Illustration of the Persona Branch’s merging mechanism. }
    \label{fig:case1}
\end{figure*}

\begin{figure*}[h]
    \includegraphics[width=\textwidth]{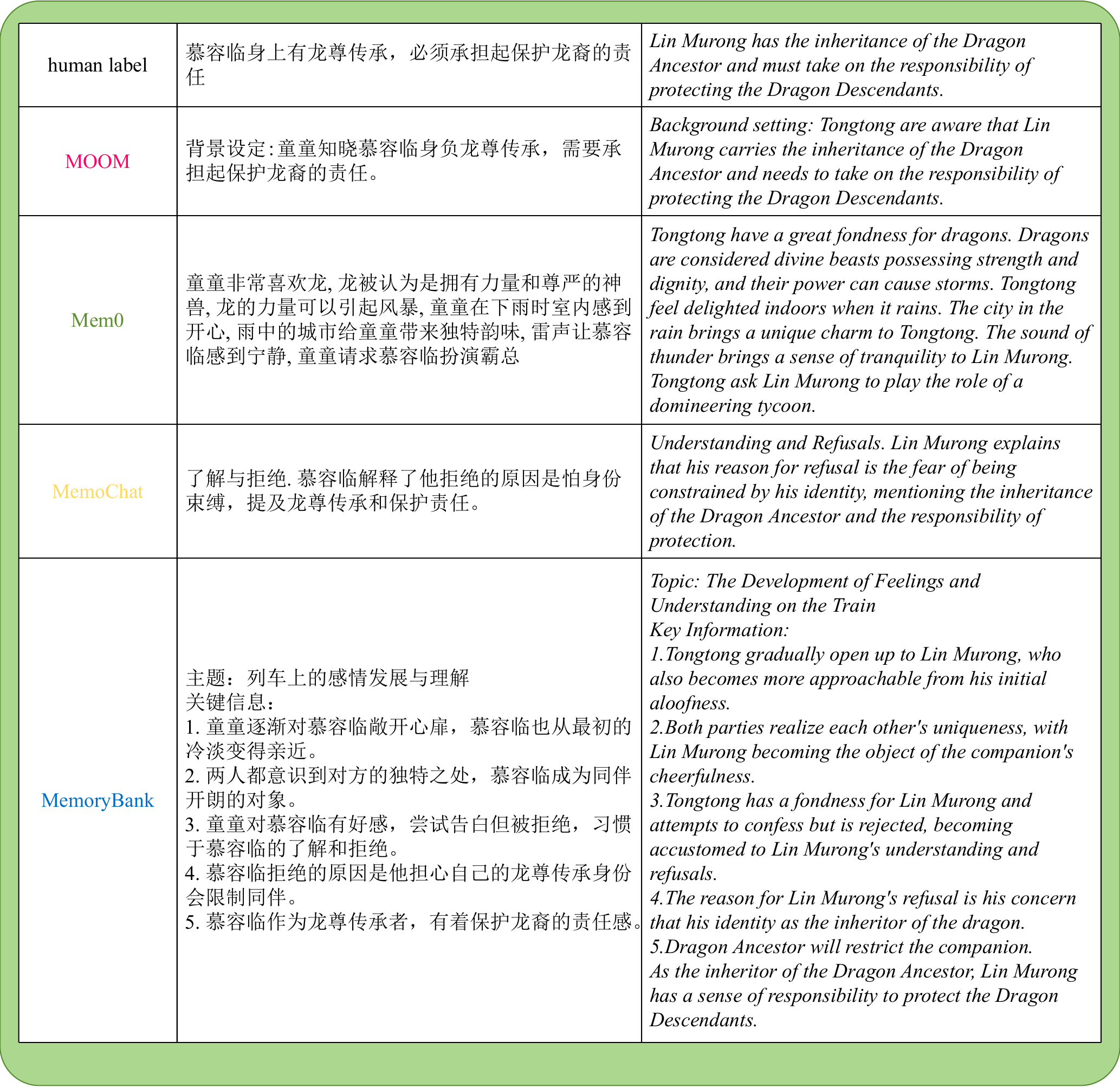}
    \caption{Illustration of the Persona Branch’s merging mechanism. }
    \label{fig:case2}
\end{figure*}

This section presents two examples comparing different frameworks in retrieving memory entries, as shown in Fig \ref{fig:case1} and Fig \ref{fig:case2}. Each example includes a label, its most similar memory information (ranked by BGE similarity) in each framework, and their English translations. The results show that Mem0 and MemoChat often lose critical details, leading to incomplete memory retrieval. In contrast, MemoryBank retains complete information but frequently consolidates excessive details into a single memory entry, risking information redundancy that interferes with the LLM’s responses. While MemoryBank performs similarly to MOOM on metrics like BERTScore and M3E, its lower performance in probe-based Q\&A tests highlights the drawback of overloading memory. MOOM’s concise and focused memory snippets effectively balance clarity and completeness, ensuring accurate and efficient question answering.

\end{document}